\documentclass[letterpaper, 10 pt, conference]{ieeeconf}
\usepackage[utf8]{inputenc}

\IEEEoverridecommandlockouts    
\usepackage[style=ieee]{biblatex}
\usepackage[english]{babel} 
\usepackage{amssymb}
\usepackage{amsmath}
\usepackage{graphicx}

\graphicspath{ {./images/} }
\usepackage{float}
\usepackage{comment}
\usepackage{booktabs}
\title{\Large \bf 
A First-Order Approach to Model Simultaneous Control of Multiple Microrobots}
\author{Logan E. Beaver, \emph{IEEE Student Member}, Sambeeta Das, \emph{IEEE Member},\\ Andreas A. Malikopoulos, \emph{IEEE Senior Member}%
\thanks{The authors are with the Department of Mechanical Engineering at the University of Delaware, Newark, DE, 19716, USA. Emails: \{lebeaver,samdas,andreas\} @ udel.edu. } }

\begin{document}
\maketitle

\begin{abstract}

The control of swarm systems is relatively well understood for simple robotic platforms at the macro scale. However, there are still several unanswered questions about how similar results can be achieved for microrobots.
In this paper, we propose a modeling framework based on a dynamic model of magnetized self-propelling Janus microrobots under a global magnetic field. We verify our model experimentally and provide methods that can aim at accurately describing the behavior of  microrobots while modeling their simultaneous control. The model can be generalized to other microrobotic platforms in low Reynolds number environments.

\end{abstract}

\section{Introduction}

Living organisms that are capable of remarkable large-scale organization and coordination—the kind of which seen in fish schools, vegetation patterns, or microbial mats—are hypothesized to coordinate their behavior by using a mix of short and long-distance interactions, including physical contact and chemical communication.
It is reasonably well understood how to engineer coordination in large collections of relatively simple communication-sensing-actuation platforms at the macro scale \cite{Oh2015}. However, there are many open questions about how similar behaviors can be achieved at the micro scale ($<$100 micron).
Micron scale robots have seen a lot of interest in the last three decades \cite{honda1996micro} due to their potential applications in fields such as targeted drug delivery \cite{sitti2015biomedical, troccaz2008development}, cleaning clogged arteries, cell sorting, biopsy \cite{Barcena2009ApplicationsBiomedicine}, cell manipulation \cite{Sakar2011WirelessMicrotransporters,Jagerpaper,kim2013fabrication,Steager2013AutomatedMicrorobots}, microsurgery \cite{Guo2007MechanismApplication}, mixing of particles \cite{peyer2013bio}, and micro-assembly \cite{Worn2000FlexibleTasks, hu2011micro}. 
Since actuators and transducers cannot be scaled down to the micron level, numerous actuation techniques have been proposed for microrobots, including electrophoretic actuation \cite{kim2015electric}, optical actuation \cite{palima2013gearing}, magnetic field actuation \cite{chowdhury2016towards}, thermal actuation \cite{ErdemThermallyMicrorobot}, and by attachment to swimming microorganisms \cite{Behkam2007BacteriaMicrorobots}.
In spite of this progress, there are several challenges associated with controlling multiple microrobots simultaneously at low Reynolds numbers.
First, microscopic, or kinematic, reversibility means that actions at the microscale are time-independent. 
Second, fluid motion at the micro and nanoscale imparts Brownian motion, and thus the trajectory of any driven robot in a solution tends to become stochastic over measurable intervals of time due to Brownian diffusion \cite{das2018experiments,vinagre2016there}. 
Third, controlling multiple magnetic microrobots simultaneously presents a challenge.
This comes from the fact that all the microrobots receive the same global control signal, which couples their dynamics. 
Finally, there is either a repulsive or attractive force between magnetic microrobots due to the magnetic field gradients generated by each individual. 
This attraction and repulsion can be manipulated to navigate the individual microrobots by controlling the orientation of the magnets with the external field \cite{Salehizadeh2017Two-agentDimensions}.
Methods that use selective electrostatic clamps \cite{chowdhury2016towards,Go2015ElectromagneticMicrorobot} or inconsistencies in the field \cite{Wong2016IndependentPlane} to navigate individual microrobots have been successfully demonstrated. However, these approaches require structured and striped surfaces and environments. 
Even with these issues resolved, interference between microrobots in a group yields a smaller net velocity than an unaccompanied microrobot \cite{Servant2015ControlledFlagella}.

Previously, ensemble control of multiple microrobots without the application of heterogeneous control schemes or specialized substrate surfaces was introduced in \cite{das2018experiments}, where we combined magnetic control with catalytically powered microrobots.
In particular, we designed microrobots that were powered by chemical reactions \cite{DasBoundariesSpheres}, and were steered by a global magnetic field.
These microrobotic systems utilize commercially available paramagnetic colloids and overlay their inherent heterogeneous magnetizations with the restrictions of catalytic self-propelled motion.

\subsection{Motivation}

While there have been a number of control algorithms for multi-agent robot systems \cite{rossi2021multiagent}, a good model of microrobot motion is necessary before model-based control algorithms can be developed.
The motion of these microrobots depends on a number of microscale forces, e.g., surface tension, surface friction, and viscous forces. An accurate estimation of these forces is very challenging. For example, in our previous work \cite{das2018experiments}, we attempted to model the behavior of the catalytic magnetic microrobots using a simple kinematic model.
This kinematic model had only two parameters: the magnetic field strength, which was a binary on-off signal, and the initial speed of the microrobots.
However, this kinematic model was insufficient to model the simultaneous control of multiple microrobots 
as it did not accurately capture the dynamics of the microrobots and the associated acting forces.
In this paper, we introduce a stokes flow based mathematical model 
that accurately models these multifunctional catalytic and magnetic microrobots.
We also demonstrate that this model accurately describes the simultaneous control of multiple catalytic and magnetic microrobots. 

These microrobots are different from the purely phoretic swimmers in \cite{GolestanianDesigningNano-swimmers}, since they have a magnetic component along with the catalytic reaction driving their motion.
The magnetic field is designed to be weak and not drive the forward motion of the microrobots, instead it act as a steering knob to control the orientation of multiple microbots simultaneously. 
In addition, these microrobots are confined to 2D due to their hydrodynamic interactions 
with the environment, as described in \cite{DasBoundariesSpheres,das2018experiments}.
Using our proposed physics-based model, 
we are able to match the simulated behavior of a microrobot experiment in simulation. 

This remainder of the paper is structured as follows.
In Section \ref{sec:system}, we describe the experimental design for the catalytic microrobots and demonstrate the previous results. 
In Section \ref{sec:modeling}, 
we provide a mathematical model of microrobot motion. In Section \ref{sec:model}, 
we demonstrate our modeling approach and how it pertains to the experiments. 
Finally, in Section \ref{sec:conclusions}, we 
delineate the conclusions and directions for future research in this domain.

\section{System Description and Experimental Results} \label{sec:system}

Catalytic microrobots \cite{Sia2003MicrofluidicStudies} self propel at the microscale using the decomposition of hydrogen peroxide (H${}_{2}$O${}_{2}$) by platinum into water and oxygen. 
This mode of propulsion has become fairly popular in microrobotics, with a multitude of catalytic powered microrobots being developed \cite{Kline2005CatalyticNanorods,Mirkovic2010NanolocomotionNanorotors,das2018experiments}. 
Their thrust mechanism is self-electrophoretic by nature \cite{Brown2015SwimmingCrystal}.
Self-propelled microrobots are appealing devices for various microscale tasks, as they are capable of fully autonomous motion \cite{sitti2015biomedical}.
Our magnetic Janus microrobots consist of a non-catalytic hemisphere and a platinum-covered hemisphere, which provide the particle with a propulsion force due to catalytic decomposition of the hydrogen peroxide on the platinum surface.
The Janus microrobots also incorporate a magnetic component in the form of embedded magnetic particles, which allow the microrobot to align its magnetic moment along the direction of an external magnetic field \cite{Baraban2012CatalyticDelivery}.
Incorporating these magnetic particles enables the orientation of all Janus microrobots to be controlled simultaneously through an external magnetic field.

Our system consists of $4.6$ $\mu $m spherical polystyrene particles doped with magnetic nanoparticles. 
These catalytic Janus spheres are prepared by spin coating a $0.1$\% (by weight)  dispersion of paramagnetic polystyrene microspheres (Spherotech, diameter = $4.6$ $\mu m$) from ethanol onto freshly cleaned glass microscope slides. A $10$ nm thick layer of platinum ($>99.9$\% Sigma Aldrich) is then evaporated onto one side of the microspheres under vacuum in a Lesker PVD75 electron beam evaporator. The magnetic field is generated using four identical in-plane electromagnetic coils.

In our earlier work \cite{das2018experiments}, we demonstrated simultaneous steering of multiple Janus microrobots to arbitrary locations using four, in-plane, electromagnetic coils.
The Janus microrobots were contained in a solution of  H$_2$O$_2$  within a sealed chamber in the center of the setup.
Each microrobot’s forward velocity was controlled by catalysis, resulting in motion directed along the axis that passes through the center of the catalytic cap. 
The microrobots' heading was controlled via magnetization. Since the magnetic microrobots have different directions of magnetization, relative to the hemisphere of the catalytic caps, we obtained different directions of motion from a uniform global magnetic field. 
We also demonstrated open loop control of multiple microrobots using a single uniform global magnetic field (see supplementary \textbf{video 1}), and were able to move the microrobots in three different directions simultaneously.

\section{Mathematical Modeling} \label{sec:modeling}

The control of multiple Janus microrobots is challenging, as it is underactuated and operates under a single global magnetic signal.
Thus, the development of a simple yet accurate analytical model is critical to develop novel model-based control methods.
Motivated by previous work and the geometry of the experimental setup \cite{das2018experiments}, we model the particles as rigid bodies in the $\mathbb{R}^2$ plane.

\subsection{Equations of Motion} 

We express the motion of the particle using Newtonian dynamics, with a fixed workspace reference (x${}_{o}$, y${}_{o}$, z${}_{o}$), to describe the motion of a magnetic Janus microrobot in a viscous fluid,
\begin{align}
\mathbf{F}_{m}(t) + \mathbf{F}_{hd}\big(\mathbf{v}(t)\big) + \mathbf{F}_{prop}(\theta) + \mathbf{F}_{e}(t) &= m\frac{d\mathbf{v}(t)}{dt}, \label{eq:bigLinear}\\
T_{m}(t) + T_{hd}\big(\omega(t)\big) + T_{prop} + T_{e}(t) &= J\frac{d \omega(t) }{dt}, \label{eq:bigTorque}
\end{align}
where $t\in \mathbb{R}_{>0}$ is time, $\mathbf{F}_{m}(t) \in \mathbb{R}^2$ and $T_{m}(t)\in\mathbb{R}$ are the force and torque imposed on the microrobot by the magnetic field, respectively; $\mathbf{F}_{hd}(t)$ and $T_{hd}(t)$ represent the hydrodynamic drag force and torque, respectively; $\mathbf{F}_{prop}$ and $T_{prop}$ are related to the self-propulsion mechanism; $\mathbf{F}_{e}$ and $T_{e}$ correspond to the other external forces and torques;  $m$ and $J$ are the mass and the moment of inertia of the particle, respectively; $\mathbf{v}$ and $\omega$ are linear and angular velocities of the particle, respectively.

\subsection{External Magnetic Control}

The particle with magnetization $\mathbf{M}$ experiences a force and torque when in the presence of a magnetic field, $\mathbf{B}$. 
The Biot-Savart equation describes the magnetic field engendered by a current loop,
\begin{align}
\mathbf{B} = \frac{\mu_0}{4}\int_{C} \frac{I(t)\,\hat{l}}{|l|^2} ds,
\end{align}
where $\mu_o$ is the permeability constant for air, $I(t)$ is the current passing through the loop of wire, $\hat{l}$ is the unit vector from the wire segment to a point of focus in the workspace, and a closed-loop integral is taken around the entire current loop, $C$, for each wire segment $ds$.

The underlying principle for the navigational control of the microrobots is to control the magnetic field, $B$, in order to induce the magnetic force and accompanying torque on the magnetized particle operating in the $\mathbb{R}^2$ plane. 
The magnetic force 
and torque 
generated by a magnetic flux $\mathbf{B}$ is
\begin{align}
\mathbf{F}_m(t) &= V\,(\mathbf{M}\cdot\nabla)\,\mathbf{B}(t), \label{eq:magnetThrust}\\
\mathbf{T}_m(t) &= V \, ( \mathbf{M} \times \mathbf{B}(t) ) \cdot \hat{\mathbf{z}}, \label{eq:magnetTorque}
\end{align}
where $V$ is the volume of the robot, $\mathbf{M}$ is the magnetization of the robot, and $\hat{\mathbf{z}}$ is a unit vector pointing out of the $\mathbb{R}^2$ plane.
In practice, the externally applied magnetic fields are designed to engender minor forces on the microrobots and substantial torques. 
Hence, magnetic forces are negligible relative to catalytic forces, yet torques cause nearly instantaneous reorientation of microrobots.
Thus, in the later derivation of our model, we will neglect the effect of \eqref{eq:magnetThrust} on microrobot motion.
The value for the magnetization of the microrobot, $\mathbf{M}$, depends on the geometry and material properties of the microrobot.

\subsection{Hydrodynamic Interactions}

In our analysis, we use the Navier-Stokes equations to model the hydrodynamics of a rigid body moving through a fluid.
Assuming a steady, incompressible viscous fluid with low Reynolds number, the drag force, ${\boldsymbol{\mathrm{F}}}_{hd}$, and torque, ${\boldsymbol{\mathrm{T}}}_{hd}$, on a spherical Janus microrobot \cite{Soc.OnPendulums} are given by
\begin{align}
\mathbf{F}_{hd}\big(v(t)\big) &= - 6\,\pi\,\eta\, r\, \mathbf{v}(t) \label{eq:hydrodynamics} \\
T_{hd}\big(\omega(t)\big) &= -8\,\pi\,\eta\, r^3\, \omega(t) , \label{eq:hydrodynamicTorque}
\end{align}
where $r$ is the radius and velocity of the particle and $\eta $ is the dynamic viscosity of the fluid medium.

\subsection{Catalytic Propulsion}

The magnetic Janus microrobots 
are covered with platinum on one half and swim in 5-20\% H${}_{2}$O${}_{2}$ solution.
They produce forward momentum via the catalytic conversion of hydrogen peroxide into water and oxygen by platinum, which gives rise to protons. Due to the unevenness of the platinum cap thickness, an ionic gradient is generated on the platinum cap between the poles and equator. 
The resulting current drives each microrobot, and they swim with their polystyrene side forward \cite{das2018experiments,ebbens2011direct}. 
For all such catalytic microrobots, their self-propulsion occurs with no external energy supply, and their thrust mechanism is self-electrophoretic by nature \cite{Brown2015SwimmingCrystal}. 
The terminal velocity of a Janus particle is highly dependent on its size \cite{ebbens2012size}, and the self-propelling force that acts on it is proportional to the concentration of H${}_{2}$O${}_{2}$ in the environment \cite{DasBoundariesSpheres}, i.e.,
\begin{align}
    \mathbf{F}_{prop}(\theta)  \propto  C(\text{H}_2\text{O}_2) \, R_{\theta} \hat{\mathbf{x}},
\end{align}
where $C($H$_2$O$_2)$ is the concentration of H${}_{2}$O${}_{2}$, and $R_{\theta}$ describes the rotation of the microrobot's polystyrene face relative to the reference axis, $\hat{\mathbf{x}}$.
The effect of catalytic propulsion is negligible, and thus we let $T_m = 0$ for our analysis.


\subsection{External Disturbances}

The external disturbance force 
and torque 
terms include any relevant external nano-scale forces, e.g., electrostatic, van der Waals, and thermal actions.
In this paper, the motion of a microrobot is assumed to be disturbed only by random fluctuations, which we consider to be Brownian.
The Langevin equation is frequently used to model the Brownian motion is $\mathbf{F}_e(t) ={}^f_{\ }\boldsymbol{\xi}(t)$ , $\mathbf{T}_e(t) ={}^t_{\ }\boldsymbol{\xi}(t)$,
where ${}^f_{\ }\boldsymbol{{\xi}}(t)$ and ${}^t_{\ }\boldsymbol{{\xi}}(t)$ represent the stochastic force and torque due to random fluctuations, respectively.

\section{Proposed Physics-Based Model} \label{sec:model}

In this section, we derive a simplified physics-based model of the micro-robot using standard linear control techniques \cite{ControlsBook}.
We take take the expectation of \eqref{eq:bigLinear} and \eqref{eq:bigTorque} with respect to $\mathbf{F}_e$ and $T_E$, respectively, and we neglect the force imposed on the Janus particle as it moves through the magnetic field.
For \eqref{eq:bigLinear}, this yields a deterministic equation that captures the interaction between the self-propelling force and the hydrodynamic drag, i.e.,
\begin{equation} \label{eq:freeBody}
    m\dot{\mathbf{v}}(t) = F R_{\phi}\mathbf{u}(t) - 6\pi\eta r \mathbf{v}(t),
\end{equation}
where $m$ is the mass of the particle, $F= |\mathbf{F}_{prop}(\theta)|$ is the magnitude of the self-propelling force, $\mathbf{u}(t)$ is a unit vector pointing in the direction of the magnetic field, $R_{\phi}$ is a rotation matrix that captures the offset angle between the particle's direction of motion and the magnetic field, and the remaining hydrodynamic terms come from \eqref{eq:hydrodynamics}.
Note that the Brownian motion terms do not appear in \eqref{eq:freeBody}, as it has zero-mean.
Several parameters of \eqref{eq:freeBody} have standard values, and to determine the magnitude of the self-propelling acceleration we extracted position and velocity data from a portion of the experimental data presented in \cite{das2018experiments} 
(see supplementary \textbf{video 1}).
Using this data, we estimated the terminal velocity of each particle and applied \eqref{eq:freeBody}, i.e., let $\dot{\mathbf{v}}(t) = 0$, to determine the value of $F$ for each particle.
These values are presented in Table \ref{tab:parameters}.

\begin{table}[ht]
    \centering
    \caption{Self-propelled particle data fit the experiment, data and the standard values for each system parameter.}    
    \begin{tabular}{c|cccc}
        F/m (m/s$^2$) & m (ng) & $\eta$ (cP) & $R$ ($\mu$m)  \\ \toprule
        1.00, 1.18, 1.34 & 0.401 & 1.245 & 4.6 
    \end{tabular}
    \label{tab:parameters}
\end{table}

Rearranging \eqref{eq:freeBody} yields a first-order linear dynamical system for the velocity of the microrobot in the direction of $R_{\phi}\mathbf{u}(t)$,
\begin{equation} \label{eq:dynamics}
    \dot{v}_{\phi}(t) + \frac{6\pi\eta r}{m} v_{\phi}(t) = \frac{F}{m} ,
\end{equation}
where $v_{\phi}(t) = \mathbf{v}(t)\cdot R_{\phi} \mathbf{u}(t)$.
The dynamics in \eqref{eq:dynamics} correspond to a first-order linear system with a constant forcing function of $\frac{F}{m}$.
Substituting the values in Table \ref{tab:parameters} yields a time constant of $\tau = 3.72$ $\mu$s, which is six orders of magnitude smaller than the timescale of microrobotic experiments.
This also implies that the microparticle achieves $95$\% of its terminal velocity within $3\tau = 11.16$ $\mu$s.
Next, we simplify \eqref{eq:dynamics} to only consider the steady-state velocity in the direction of motion, i.e., $\dot{\mathbf{v}}(t) = 0$ at steady state,
\begin{equation} \label{eq:vss}
    v_{ss} = v_{\phi}(t) = \frac{F}{6\pi \eta r},
\end{equation}
where $v_{ss}$ is the steady-state velocity in the direction of $R_{\phi}\mathbf{u}$.
As an example of this steady-state model's accuracy, consider that the distance a microrobot must move before the transient period is less than $1$\% of the total trajectory.
We estimate this distance by taking the product of $v_{ss} $ and $ 300\tau$, which yields, for the values in Table \ref{tab:parameters}, distances of $40.3$ nm, $47.5$ nm, and $54.0$ nm.
This equates to approximately $1\%$ of the radius of a particle, which clearly demonstrates that there is no benefit to utilizing Stokes' Drag compared to a first-order terminal velocity model.

Next, consider the rotational forces that impact the motion of a particle.
We take the expectation of \eqref{eq:bigTorque} with respect to the external disturbances and let $T_m = 0$.
The expectation yields the deterministic interaction between the torque imposed by the magnetic field \eqref{eq:magnetTorque} and hydrodynamic drag from Stokes' Law \eqref{eq:hydrodynamicTorque},
\begin{equation}\label{eq:freeBodyRotation}
    I\dot{\omega}(t) = \Big((d\,R_{\phi}R_{\theta}\hat{\mathbf{x}})\times(\mathbf{u}(t)B)\Big)\cdot\hat{\mathbf{z}} - 8\pi\eta r^3\omega(t),
\end{equation}
where $I$ is the moment of inertia of the particle, $d$ is the magnitude of the particle's dipole moment, 
$\mathbf{u}(t)B = \mathbf{B}(t)$ denotes the external magnetic field, 
and $\omega(t)$ is the angular rotation rate of the particle in the $\mathbb{R}^2$ plane.
This implies a linear first-order system with a non-linear forcing function,
\begin{equation} \label{eq:rotationDynamics}
    \dot{\omega}(t) + \frac{8\pi\eta r^3}{I}\omega(t) = \Big((R_{\phi}R_{\theta}\hat{\mathbf{x}}^T\times\mathbf{u}(t))\Big)\cdot\hat{\mathbf{z}}\frac{B\,d}{I},
\end{equation}
which, using the parameters Table \ref{tab:parameters}, yields a time constant of $\tau = 1.11$ $\mu$s.
Note that while the forcing function contains the angle of the particle, i.e., $\dot{\theta}(t) = \omega(t)$, it is equal to zero when the particle's dipole is aligned with $\mathbf{u}(t)$. 
Additionally, the dynamics of \eqref{eq:rotationDynamics} drives the forcing function to zero.
Therefore, we expect $\omega(t)$ to decay to zero, and the particle to align with the magnetic field, with the same order of magnitude as the time constant.

Thus, as with the linear motion, we expect the particle to align with the magnetic field on a timescale that is orders of magnitude faster than the timescale of motion control.
This leads to our final first-order dynamic model for motion of a Janus particle,
\begin{align} \label{eq:finalDynamics}
    \dot{\mathbf{p}}(t) = v_{ss} \mathbf{R}_{\phi} \mathbf{u}(t).
\end{align}
Our analysis shows that our proposed model will have a negligible variation from a Stokes' drag-based model, and in the following section we show that it incurs a lower computational cost.
Finally, note that while our model is not stochastic, one could argue that it is impossible to predict the exact values that the Brownian motion will take in a physical experiment.
Therefore, having a computationally fast microrobot model will allow real-time feedback control techniques that compensate for particle drift throughout the experiment.

\section{Experimental Validation} \label{sec:experiment}

To validate our model, we simulated the 3-particle experiment from \cite{das2018experiments} using our proposed model and compared it to the full dynamic model using Stokes' law.
In this experiment, each particle moves in a straight line for $2$--$6$ seconds while the control input (magnetic field direction) remains constant.
We extracted the initial position of the particles from a video recording of the experiment and estimated the terminal velocity of each particle over one time interval of constant control input to derive the values of $F$ in Table \ref{tab:parameters}. 
Next, to determine the value of $\phi$ for each particle, we applied a linear regression over the same interval.
These parameters are listed in Table \ref{tab:experiment} along with the experimental results, which we discuss next.

\begin{table}[ht]
    \centering
    \caption{Particle angle and resulting root mean square error (RMSE) for the simulated particles to track the experiment.}
    \begin{tabular}{cccc}
                   & $\phi$         & RMSE (Approach 1) & RMSE (Approach 2) \\\toprule
        Particle 1 &  -1.09 rad  &  4.11 $\mu$m &  6.91 $\mu$m \\
        Particle 2 &   3.82 rad  &  5.62 $\mu$m &  5.56 $\mu$m \\
        Particle 3 &   2.64 rad  &  4.02 $\mu$m &  5.77 $\mu$m 
    \end{tabular}
    \label{tab:experiment}
\end{table}

First, to demonstrate the benefits of our approach over the full model, we performed each of the simulations using Stokes' law \eqref{eq:dynamics} and our proposed model \eqref{eq:vss} to compare their accuracy and computational cost on a desktop PC (intel i5-3570, 12 GB Ram).
We simulated the microrobots in Matlab R2018b using a standard stiff numerical solver (ODE15s) for the Stokes' law model and a forward Euler's method with a time step of approximately $0.02$ seconds for our proposed model.
Table \ref{tab:computational} presents the results of these experiments, where the computational time is averaged over $100$ runs and the RMSE is relative to the experimental data.
Note that the actual time of the physical experiment is approximately $15$ seconds.
Thus, while each model may be sufficiently fast to embed in a model-based control framework, our approach has the benefit of running  $5$\% faster and having no discernible impact on the overall accuracy of the simulation.
Furthermore, our model is first-order and is therefore significantly simpler to embed in existing model-based control frameworks.

\begin{table}[ht]
    \centering
    \caption{Computation time comparison for Stokes' Law and our proposed model.}    
    \begin{tabular}{ccc}
            & Stokes' Law  & Proposed  \\\toprule
         Runtime:  & $2.13$ seconds & $2.04$ seconds \\
         Mean RMSE : &  $4.59$ microns  & $4.59$ microns
    \end{tabular}
    \label{tab:computational}
\end{table}

\begin{figure}[ht]
    \centering
    \includegraphics[width=\linewidth]{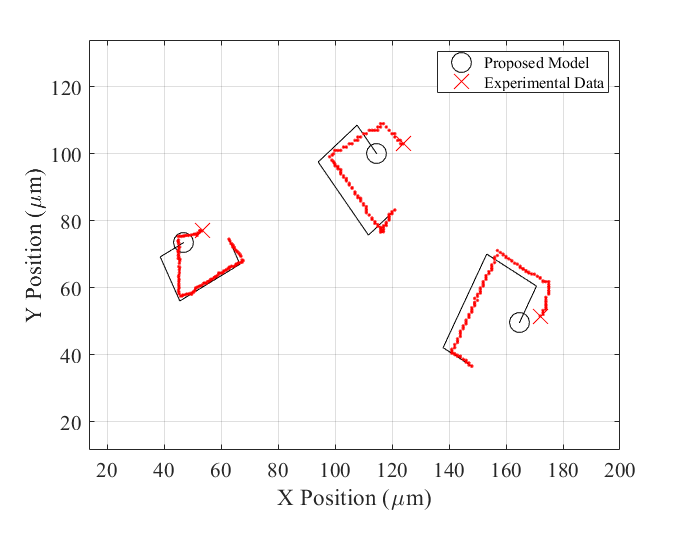}
    \caption{Comparison of experimental data and our simulated model, where each particle's terminal velocity was extracted from a portion of the data.}    \label{fig:many-F}
\end{figure}

Next, we performed two simulations to quantify the effect of the particle's terminal velocity on the accuracy of the simulation. 
In the first approach, used the values of $\frac{F}{m}$ in Table \ref{tab:parameters}.
The trajectories of each particle for this case are presented in Fig. \ref{fig:many-F} overlaid on the experimental data.
Each of the particles is able to track the longitudinal position of the experimental particle within a few microns, although in each case the experimental particles experience lateral drift relative to their simulated counterparts.
However, we expect this drift to emerge in any model of the microrobots' stochastic motion.
To quantify the error between the physical and simulated particles, we calculated the root mean square error (RMSE) between the simulated and actual position at each time step, using the experiment clock.
This value is reported in Table \ref{tab:experiment} for both approaches.
Due to Brownian motion, the tracking error does increase with time, however, the RMSE at the end of the experiment is not significantly larger than a single particle's radius ($4.6$ $\mu$m).

\begin{figure}[ht]
    \centering
    \includegraphics[width=\linewidth]{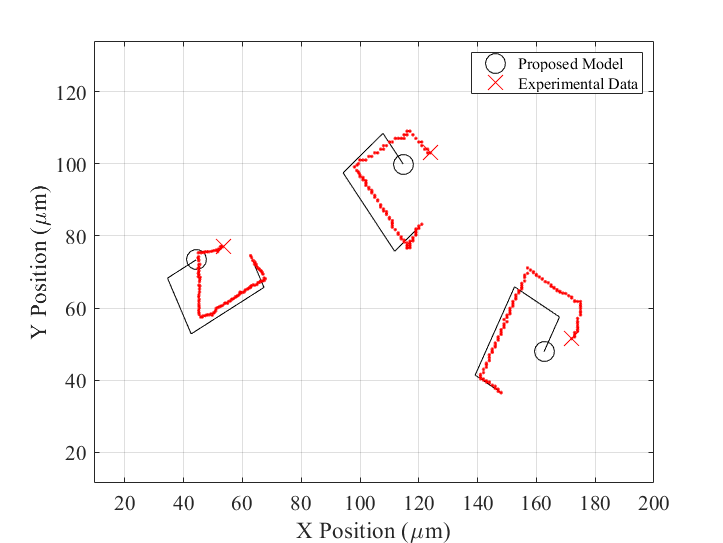}
    \caption{Comparison of experimental data and our simulated model, where all particles share a common terminal velocity.}
    \label{fig:one-F}
\end{figure}

In our second approach, we selected the mean value of $\frac{F}{m} = 1.17$ m/s$^2$ from Table \ref{tab:parameters} and gave each particle an identical terminal velocity. 
This approach may be more representative of our model's performance, as the value of $F$ may not be known a priori.
The trajectories of each particle are presented in Fig. \ref{fig:one-F} overlaid on the experimental data.
This case leads to significantly more error, as only the median particle with $F = 1.18$ (top-center of Fig. \ref{fig:one-F}) is physically similar to the imposed terminal velocity. 
This is confirmed in Table \ref{tab:experiment}, where Particles 1 and 3 (left and right side of Fig. \ref{fig:one-F} respectively) have a higher RMSE tracking error.
In fact, the average difference in RMSE between the first and second approach is approximately $30$\%, which coincides with the the $30$\% average deviation in the value of $F$ in Table \ref{tab:parameters}.
These observations make us feel comfortable argue that the model is capturing the behavior of the particles, and that the RMSE is directly proportional to the error in estimating system parameters.

\section{Conclusions} \label{sec:conclusions}

In this paper, we demonstrated that a first-order physics-based model of a magnetized self-propelled particle can accurately describe simultaneous control of catalytic microrobots controlled by a single external magnetic signal.
The equations describing the microrobot dynamics can be generalized and used for most microrobots in low Reynolds number systems.
In our exposition, first, we adapted a deterministic physics-based model based on Stokes' Law and rigid body dynamics for these microrobots.
Then, using classical linear control techniques, we demonstrated that time scale of particle motion is separable from the hydrodynamic and electromagnetic forces, and thus their dynamics can be separated.
In addition, we showed that the proposed model is significantly faster than Stokes' Law to compute trajectories.
Finally, we validated the model by comparing the generated trajectories to experimental data.

It is our hope that this approach could potentially allow us to develop accurate motion planning and control algorithms for these microrobots in the future.
The biggest challenge in the proposed model is calculating the two parameters, i.e., the net dipole moment orientation, $\phi$, and self-propelled particle force, $F$, for each particle during an experiment.
One potential direction for future research is to develop an adaptive model-based control algorithm that can estunate these parameters online.
Analyzing the controllability of an arbitrary number of magnetically-actuated microrobots could be another compelling direction for future research along with generating particular desired formation for collective transport and drug delivery. 
Finally, including a control term that captures the deflection of particles in the presence of a heated laser point is another area that deserves futher investigation. 


\printbibliography

\end{document}